\documentclass[11pt]{article}

\usepackage[final]{acl}

\usepackage{times}
\usepackage{latexsym}

\usepackage[T1]{fontenc}
\usepackage[utf8]{inputenc}
\usepackage{microtype}
\usepackage{inconsolata}

\usepackage{graphicx}
\usepackage{booktabs}
\usepackage{multirow}
\usepackage{amsmath}
\usepackage{amssymb}

\usepackage{subcaption}

\graphicspath{{figures/}}

\title{Where Vision Becomes Text: Locating the OCR Routing Bottleneck in Vision-Language Models}

\author{Jonathan Steinberg$^1$ \and Oren Gal$^1$ \\
  $^1$Swarms \& AI Lab (SAIL), University of Haifa \\
  \texttt{jsteinber@staff.haifa.ac.il}}

\begin{document}
\maketitle

\begin{abstract}
Vision-language models (VLMs) can read text from images, but where does this optical character recognition (OCR) information enter the language processing stream? We investigate the OCR routing mechanism across three architecture families (Qwen3-VL, Phi-4, InternVL3.5) using causal interventions on five dense models in the 2B--8B parameter regime. By computing activation differences between original images and text-inpainted versions, we identify \textbf{architecture-specific OCR depth bands} whose dominant location depends on the vision-language integration strategy: DeepStack models (Qwen) show peak sensitivity at mid-depth ($\sim$50\%) for scene text, while single-stage projection models (Phi-4, InternVL) peak at early layers (6--25\%), though the exact layer of maximum effect varies across datasets. The OCR signal is remarkably low-dimensional: PC1 captures up to 72.9\% of variance. Crucially, Principal Component Analysis (PCA) directions learned on one dataset transfer to others, demonstrating shared text-processing pathways. Surprisingly, in models with modular OCR circuits (notably Qwen3-VL-4B), OCR removal can \emph{improve} counting performance (up to +6.9pp), suggesting OCR interferes with other visual processing in sufficiently modular architectures within this model-size regime. Beyond interpretability, our method provides a diagnostic framework for assessing VLM vulnerability to visual prompt injection attacks, where adversarial text embedded in images can hijack model behavior.
\end{abstract}

\section{Introduction}

Vision-language models (VLMs) have become increasingly capable at reading text from images, enabling applications from document understanding to visual question answering. This optical character recognition (OCR) capability emerges from training on web-scale data, but the internal mechanisms remain poorly understood.
\\\\
Understanding \emph{where} OCR information enters the language processing stream is crucial for both interpretability and safety. Unlike explicit text inputs, OCR-derived text must be extracted from visual representations and routed into the residual stream where it can influence generation. Localizing this routing mechanism enables targeted interventions for understanding and controlling OCR-based processing.
\\\\
Beyond interpretability, this question has direct safety implications. As VLMs are increasingly deployed in agentic settings---from robotic control to autonomous web browsing---adversarial text embedded in images (typographic prompt injections) can hijack model behavior \citep{westerhoff2025scam}. Understanding the internal pathway through which image-embedded text influences model outputs is a prerequisite for mechanistic defenses against such attacks. In this work, we investigate the OCR routing bottleneck across three VLM architecture families using PCA-based causal interventions. 
\\
Our contributions are:

\begin{figure*}[t]
    \centering
    \includegraphics[width=\textwidth]{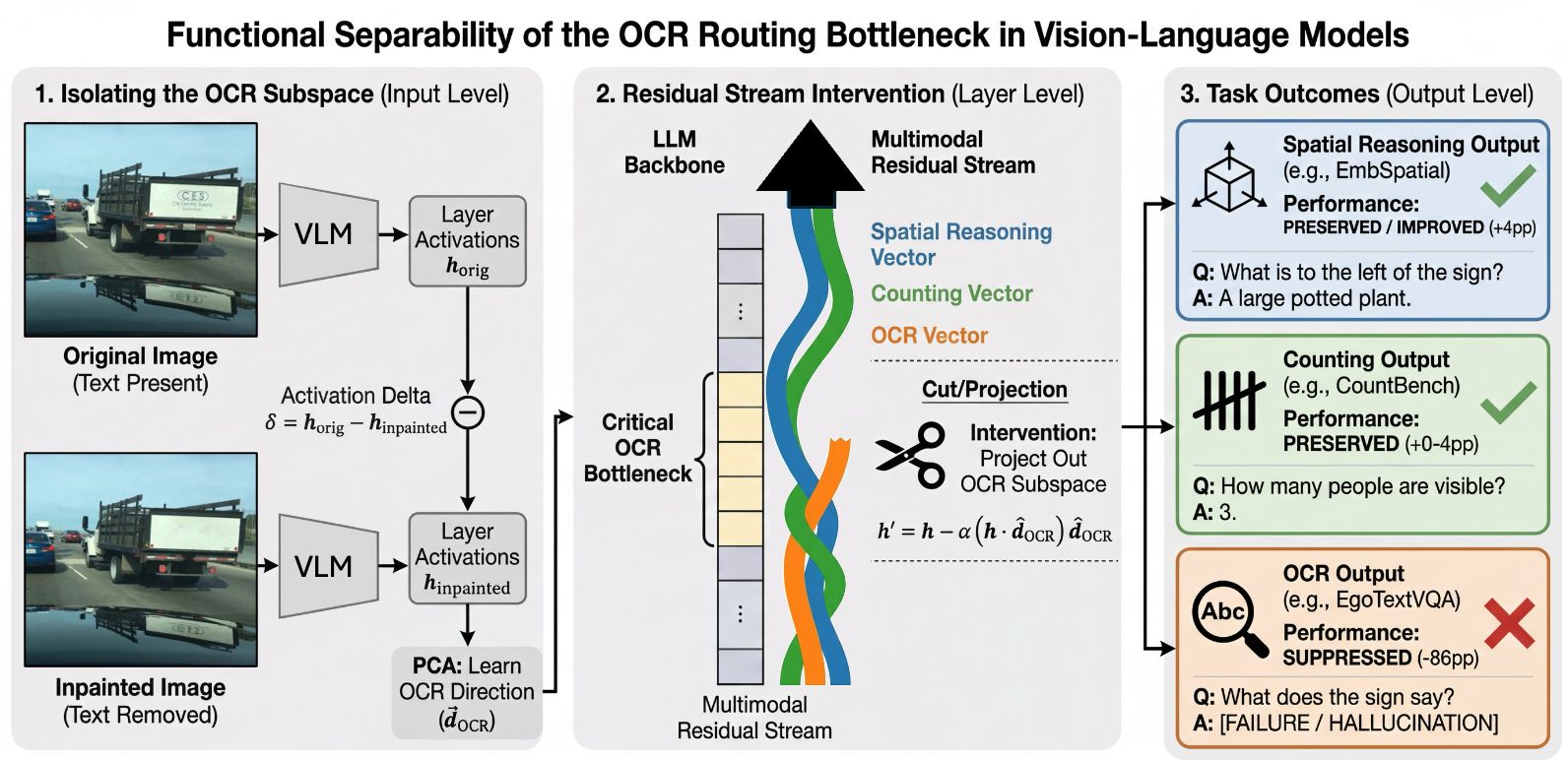}
    \caption{\textbf{Causal intervention on the VLM residual stream.} By identifying the OCR direction via PCA on activation deltas (left) and projecting it out at the bottleneck (center), OCR capability is effectively suppressed while spatial reasoning and counting tasks are functionally separated and preserved (right).}
    \label{fig:method_overview}
\end{figure*}
\begin{enumerate}
    \item We identify \textbf{architecture-specific OCR depth bands} where OCR becomes causally usable, replicated across five dense models (2B--8B) from three architecture families (Qwen3-VL, Phi-4, InternVL3.5-4B), with the dominant sensitivity regime determined by the vision-language integration strategy.
    \item We show the OCR signal is \textbf{remarkably low-dimensional}: PC1 explains 72.9\% of variance, enabling efficient subspace removal.
    \item We demonstrate \textbf{cross-dataset transfer}: PCA directions learned on EgoTextVQA transfer to OCRBench and InfoVQA, indicating shared text-processing pathways.
    \item We discover a \textbf{synergistic improvement} in Qwen3-VL-4B: the L17--19\_pc5 intervention improves counting (+4.3pp) and general VQA (+1pp) simultaneously with only mild spatial degradation ($-$2pp); the peak counting gain reaches +6.9pp (L16--20\_pc3), suggesting OCR interferes with visual processing in models with sufficiently modular circuits within the evaluated 2B--8B dense model regime.
\end{enumerate}

\section{Related Work}

\paragraph{Vision-Language Model Architectures.}
Modern VLMs integrate visual and textual information through various fusion strategies. Early approaches like Flamingo \citep{alayrac2022flamingo} used cross-attention layers to inject visual features into frozen language models. LLaVA \citep{liu2023visual} introduced visual instruction tuning with a simpler projection-based integration. Qwen3-VL \citep{bai2025qwen3} employs the DeepStack architecture, which leverages multi-level ViT features for tighter vision-language alignment across transformer depth. This architectural diversity motivates layer-wise analysis: different integration strategies may route OCR information through different computational paths.

\paragraph{Mechanistic Interpretability of VLMs.}
Recent work has begun applying interpretability techniques to vision-language models. \citet{baek2025large} identified ``OCR heads''---specific attention heads responsible for text recognition---through activation analysis and head masking experiments. They found OCR heads are qualitatively distinct from general retrieval heads, with less sparse activation patterns. \citet{sheta2025behavioral} introduced VLM-Lens, a probing framework for diagnosing internal competences of VLMs across visual, linguistic, and cross-modal dimensions. \citet{skean2025layer} systematically analyzed layer-wise representations across language models, revealing how information transforms through depth. Our work builds on these descriptive and correlational analyses by adding \emph{causal} intervention: we identify the OCR-specific subspace in the residual stream and test whether selectively removing it disrupts text reading while preserving other capabilities, establishing not just where OCR information is encoded but whether it is separable.

\paragraph{Activation Patching and Causal Tracing.}
Activation patching \citep{meng2022locating} has proven effective for localizing specific computations in language models. The core insight is that restoring ``clean'' activations to a corrupted forward pass reveals which components are causally necessary for a behavior. We adapt this approach for multimodal settings by using inpainted images (with text removed) as the ``corrupted'' condition, enabling us to isolate OCR-specific activations. Related techniques include causal scrubbing \citep{chan2022causal} and path patching, though we focus on subspace-level rather than component-level interventions.

\paragraph{Representation Engineering and Steering Vectors.}
\citet{zou2023representation} demonstrated that model behavior can be controlled by intervening on learned directions in activation space. This ``representation engineering'' paradigm has been applied to control sentiment, truthfulness, and safety-relevant behaviors. Steering vectors \citep{turner2023steering} provide similar control through additive interventions. We apply these techniques to identify and remove OCR-related directions, testing whether low-dimensional subspace removal provides causal control over text reading without collateral damage to other capabilities.

\paragraph{Typographic Attacks and OCR Control.}
Recent work has explored mechanistic approaches to controlling OCR capabilities in vision models. \citet{hufe2026dyslexify} utilized linear probes and attention head ablation to remove OCR capabilities from CLIP models, demonstrating that text-reading circuits can be surgically targeted---a finding complementary to our subspace removal approach in generative VLMs. \citet{joseph2025steering} suppressed OCR capabilities in CLIP by identifying and dropping OCR-specific features discovered via sparse autoencoders. On the evaluation side, \citet{westerhoff2025scam} introduced a dataset and evaluation framework for studying model behavior under adversarial typographic perturbations in real-world settings. Our work extends this line of mechanistic OCR analysis from discriminative (CLIP) to generative VLMs, and from component-level ablation to representation-level subspace removal.

\paragraph{Superposition and Feature Geometry.}
\citet{elhage2022toy} introduced the concept of superposition---where neural networks represent more features than they have dimensions by encoding features in nearly-orthogonal directions. Our finding that OCR-related activation \emph{differences} concentrate in a low-dimensional subspace (PC1 explains 72.9\% of variance in original$-$inpainted deltas) is consistent with OCR being relatively separable from other features, though we note this measures variance in deltas rather than providing direct evidence about the degree of superposition. This connects to broader questions about how multimodal information is geometrically organized in shared representation spaces.

\section{Methods}

\subsection{Dataset: EgoTextVQA}

\begin{figure}[t]
    \centering
    \includegraphics[width=\columnwidth]{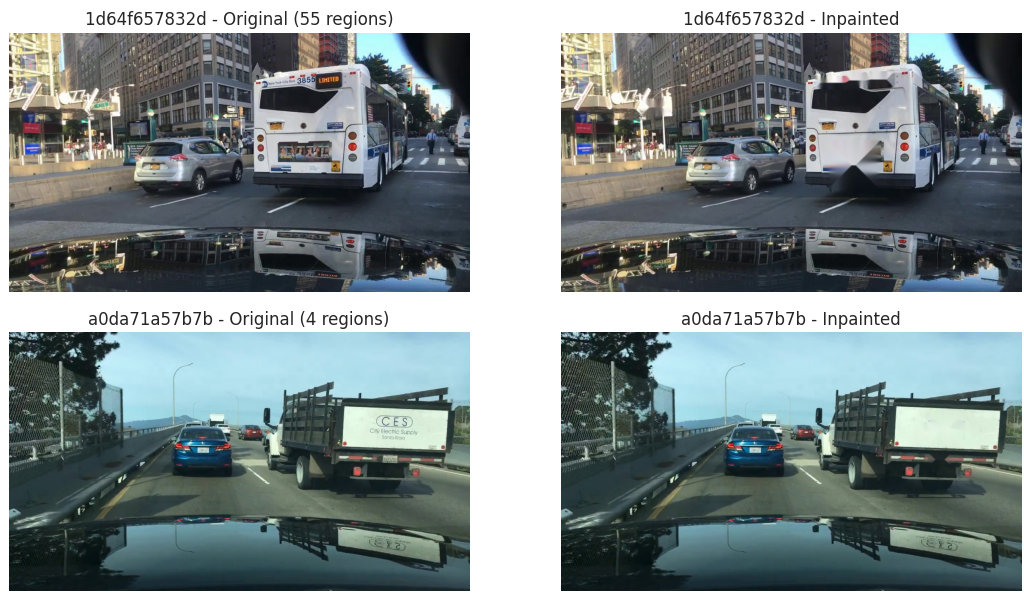}
    \caption{Example original-inpainted pairs from EgoTextVQA. Left: original images with naturally-occurring text. Right: inpainted versions with text regions removed. Activation differences between pairs capture the ``OCR signal.''}
    \label{fig:inpainting_examples}
\end{figure}

We use EgoTextVQA \citep{zhou2025egotextvqa}, an egocentric dataset where images contain naturally-occurring text (signs, labels, screens). For each image, we create an inpainted version with the text region removed (Figure~\ref{fig:inpainting_examples}). The pair (original, inpainted) enables computing what changes in activations when text is present versus absent.

\subsection{Models}

We study five VLMs spanning three architectural families to examine how vision-language integration affects OCR routing:

\begin{itemize}
    \item \textbf{DeepStack models}: Qwen3-VL-4B-Instruct, Qwen3-VL-2B-Instruct, and Qwen3-VL-8B-Instruct \citep{bai2025qwen3} progressively inject visual features into multiple LLM layers.
    \item \textbf{Single-stage models}: Phi-4-multimodal-instruct \citep{abouelenin2025phi}  and InternVL3.5-4B \citep{wang2025internvl3} inject all visual tokens at the input layer.
\end{itemize}
This architectural variation reveals how visual information integration strategy affects OCR bottleneck location and modularity.

\subsection{Cross-Dataset Benchmarks}

To validate generalization beyond EgoTextVQA, we evaluate on two additional OCR benchmarks:

\begin{itemize}
    \item \textbf{OCRBench} \citep{liu2024ocrbench}: Comprehensive OCR benchmark covering regular, irregular, and artistic text.
    \item \textbf{InfoVQA} \citep{mathew2022infographicvqa}: Document understanding benchmark with infographics and complex layouts.
\end{itemize}

\subsection{PCA-Based Subspace Removal}

For each layer $\ell$, we compute activation differences:
\begin{equation}
    \Delta h_\ell = h_\ell^{\text{original}} - h_\ell^{\text{inpainted}}
\end{equation}
across a held-out training split. We split EgoTextVQA into 315 images for PCA computation and a disjoint set for evaluation, ensuring no train-test leakage. PCA on the differences yields principal components capturing ``what changes when text is present.'' During inference, we intervene by projecting out the top $N$ components:
\begin{equation}
    h_\ell^{\text{int}} = h_\ell - \alpha \sum_{i=1}^{N} (h_\ell \cdot \text{pc}_i) \, \text{pc}_i
\end{equation}
where $h_\ell$ is the residual-stream activation at layer $\ell$, $\text{pc}_i$ is the $i$-th principal component, $N$ is the number of components removed, and $\alpha$ controls intervention strength. We fix $\alpha{=}1$ throughout (full projection removal). If OCR flows through this subspace, removal should disrupt text reading. We use the shorthand \texttt{pca\_L$\ell$-$\ell'$\_pc$N$} to denote projection of the top-$N$ components across layers $\ell$-$\ell'$.

\paragraph{Intervention Specification.}
We apply interventions via forward hooks on transformer layer outputs (post-MLP residual stream). Hooks affect all tokens (visual and text) during both prefill and decode. PCA directions are unit-normalized before projection.

\subsection{Head Ablation}

As a complementary approach, we compute a \textbf{selectivity ratio} for each attention head: the ratio of attention to OCR regions versus background regions. Heads with ratio $> 1.0$ preferentially attend to text. We rank all 720 heads (36 layers $\times$ 20 heads) by selectivity and ablate the top $N$ by zeroing their output projections during both prefill and decode, testing whether attention-identified heads are causally important.

\subsection{Evaluation}

We measure OCR accuracy using normalized substring matching (case-insensitive, punctuation-stripped), a standard evaluation metric for OCR benchmarks that counts predictions as correct if ground truth appears within the model output or vice versa.\footnote{VLM judge verification on a subset achieved 97\% agreement with this automatic metric.} To assess collateral damage, we evaluate interventions on CountBench \citep{paiss2023teaching} (object counting), EmbSpatial \citep{du2024embspatial} (spatial reasoning), and RealWorldQA \citep{realworldqa} (general visual QA) across all five models. If OCR removal hurts these tasks, the intervention is too aggressive; if it helps, OCR was interfering with visual processing. All evaluations use complete datasets with deterministic greedy decoding.

\section{Experiments and Results}

\subsection{Layerwise Bottleneck Discovery}

Figure~\ref{fig:multi_model} shows OCR accuracy across layers and intervention strengths for all five models. A clear pattern emerges across architecture families: \textbf{DeepStack models (Qwen) show mid-layer sensitivity bands}, while \textbf{single-stage models (Phi-4, InternVL) show early-layer sensitivity bands}. The bottleneck at $\sim$50\% network depth in Qwen aligns with prior findings that OCR heads concentrate at 50--60\% depth \citep{baek2025large}.

\subsection{Low-Dimensional OCR Signal}

The first principal component at L17 explains \textbf{72.9\%} of OCR-related variance. Removing PC1--PC3 achieves strong OCR suppression with fewer components than single-direction removal, confirming the signal is concentrated in a small subspace.

\subsection{Cross-Architecture Retention Analysis}

We evaluated retention across three diverse benchmarks (see Section 3.6): CountBench \citep{paiss2023teaching} (object counting), EmbSpatial \citep{du2024embspatial} (spatial reasoning), and RealWorldQA \citep{realworldqa} (general visual QA). Table~\ref{tab:cross_model_retention} reveals a striking architectural divergence.
\\\\
\textbf{Qwen3-VL-4B achieves a favorable trade-off.} Interventions at the mid-depth bottleneck (L17--19) significantly improve counting (+4.3pp) and RealWorldQA (+1pp) with only minimal impact on spatial reasoning ($-$2pp). This suggests the model has developed modular circuits where OCR can be targeted and removed, potentially freeing capacity for other visual tasks.
\\\\
\textbf{Smaller and Single-Stage Models show Trade-offs.} Qwen3-VL-2B and Phi-4 show large gains in counting (+5.6pp, +4.3pp) but at the cost of spatial reasoning and general VQA. This implies a ``competition'' for resources where removing OCR helps counting but hurts complex reasoning.
\\\\
\textbf{InternVL shows complete degradation.} Unlike other models, InternVL's early-layer bottleneck (L2--L3) degrades \emph{all} metrics when intervened. This reveals an architectural constraint: single-stage projection forces tight coupling between OCR and general vision at early layers.
\\\\
\textbf{Qwen3-VL-8B replicates the 4B pattern.} The 8B model shows a similar bottleneck at L17 with CountBench improvement (+2.9pp) and mild EmbSpatial degradation ($-$3pp). RealWorldQA shows a larger drop ($-$14pp). The L17-19\_pc5 intervention improves CountBench (+4.3pp) and RealWorldQA (+1pp) simultaneously with only mild EmbSpatial degradation ($-$2pp). L18\_pc5 causes RWQ collapse ($-$13pp) while L17/L19 preserve it.

\begin{table}[t]
    \centering
    \small
    \begin{tabular}{llccc}
        \toprule
        Model & Intervention & Count & Emb & RWQ \\
        \midrule
        Qwen3-VL-4B & baseline & 91.4 & 81.5 & 75.0 \\
         & L17-19\_pc5 & \textbf{95.7} & 79.5 & \textbf{76.0} \\
        \midrule
        Qwen3-VL-8B & baseline & 90.2 & 82.0 & 72.0 \\
         & L17-19\_pc5 & \textbf{93.1} & 79.0 & 58.0 \\
        \midrule
        Qwen3-VL-2B & baseline & 90.3 & 78.0 & 66.0 \\
         & L12\_pc5 & \textbf{95.9} & 72.0 & 60.0 \\
        \midrule
        Phi-4 & baseline & 89.2 & 62.2 & 66.0 \\
         & L8\_pc3 & \textbf{93.5} & 49.0 & 53.5 \\
        \midrule
        InternVL & baseline & 78.9 & 69.2 & 64.0 \\
         & L2\_pc1 & 69.2 & 25.0 & 42.0 \\
        \bottomrule
    \end{tabular}
    \caption{Cross-model retention results (\%). Count=CountBench, Emb=EmbSpatial, RWQ=RealWorldQA. Qwen3-VL-4B achieves favorable trade-off; smaller/single-stage models show trade-offs or degradation. Qwen3-VL-8B replicates the 4B pattern at L17.}
    \label{tab:cross_model_retention}
\end{table}

\begin{table}[t]
    \centering
    \small
    \begin{tabular}{lcccc}
        \toprule
        Intervention & Count & Emb & RWQ & Note \\
        \midrule
        \textbf{baseline} & 91.4 & 81.5 & 75.0 & --- \\
        \midrule
        \multicolumn{5}{l}{\emph{Single layer:}} \\
        pca\_L16\_pc5 & 94.1 & 81.0 & 71.0 & Count$\uparrow$, RWQ$\downarrow$ \\
        pca\_L17\_pc1 & 90.3 & 80.5 & 73.0 & Minimal \\
        pca\_L17\_pc3 & 94.1 & 78.5 & 73.0 & Trade-off \\
        pca\_L17\_pc5 & \textbf{96.2} & 78.5 & 74.0 & Good \\
        pca\_L18\_pc5 & 91.9 & 81.5 & 62.0 & RWQ collapse \\
        pca\_L19\_pc5 & \textbf{96.2} & 80.5 & 72.0 & Count$\uparrow$ \\
        pca\_L20\_pc5 & 91.4 & 79.5 & 69.0 & Trade-off \\
        \midrule
        \multicolumn{5}{l}{\emph{Multi-layer:}} \\
        pca\_L16-20\_pc1 & 89.7 & 81.0 & 65.0 & RWQ$\downarrow$ \\
        pca\_L16-20\_pc3 & 98.3 & 80.5 & 73.0 & Best Count \\
        pca\_L17-19\_pc3 & 94.1 & 78.5 & 73.0 & Good \\
        pca\_L17-19\_pc5 & \textbf{95.7} & 79.5 & \textbf{76.0} & \textbf{Best} \\
        pca\_L17-20\_pc3 & 93.5 & 79.5 & 74.0 & Good \\
        \bottomrule
    \end{tabular}
    \caption{Complete Qwen3-VL-4B retention results (\%). L17-19\_pc5 achieves the best trade-off: Count +4.3pp, RWQ +1pp, with mild EmbSpatial degradation ($-$2pp).}
    \label{tab:qwen4b_complete}
\end{table}

\subsection{Head Ablation Analysis}

As a complementary analysis, we ablated the 93 most OCR-selective attention heads (clustered in L17--24, identified by selectivity ratio $> 1.0$). This reduces OCR by 8.2pp while improving CountBench by 5.0pp---directionally consistent with our PCA findings. As a control, ablating 93 randomly-selected heads (mean over 10 draws) degrades CountBench by 6.8$\pm$2.1pp without substantially affecting OCR ($-$2.1 $\pm$ 1.5pp). Head ablation is a noisier intervention than subspace removal, but provides supporting evidence that OCR-relevant computation concentrates in the same mid-network band identified by PCA. The OCR-selective ablation is an outlier relative to random ablations: it degrades OCR while \emph{improving} counting, consistent with our interference hypothesis.

\subsection{Cross-Dataset Transfer Results}

\begin{figure*}[t]
    \centering
    \includegraphics[width=\textwidth]{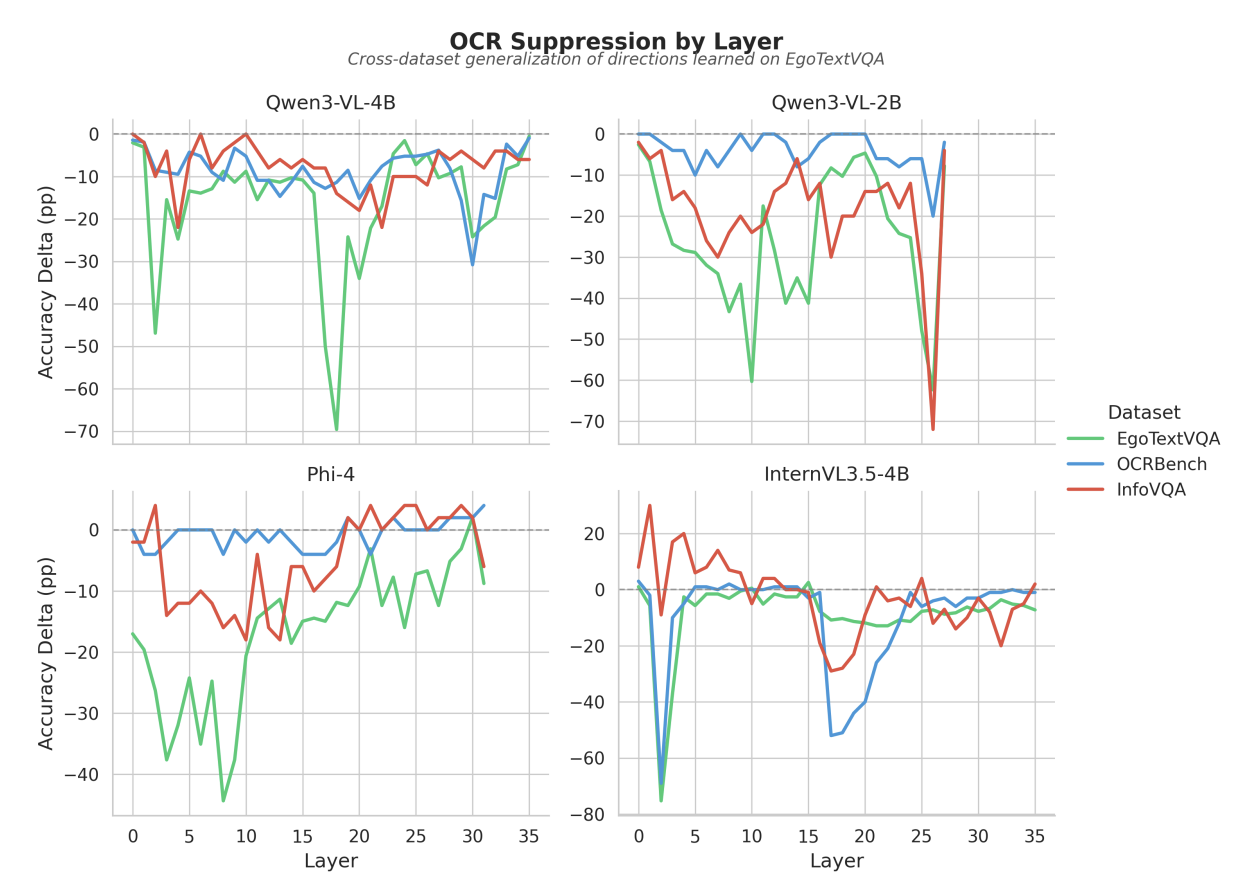}
    \caption{Cross-dataset generalization of OCR suppression. Each panel shows one model's accuracy delta across layers for all three datasets. PCA directions learned on EgoTextVQA (green) transfer effectively to OCRBench (blue) and InfoVQA (red), demonstrating that learned OCR representations generalize beyond the training distribution.}
    \label{fig:cross_dataset}
\end{figure*}

Table~\ref{tab:cross_dataset} shows maximum accuracy degradation when applying EgoTextVQA-trained PCA directions to OCRBench and InfoVQA. Critically, the PCA directions learned on one dataset effectively suppress OCR on others---demonstrating that the underlying OCR representations are shared across text domains. Note that these results test \emph{method transferability} (whether EgoTextVQA-learned directions suppress OCR on new datasets), not \emph{bottleneck transferability} (whether the max-sensitivity layer is identical). The max-sensitivity layer may shift across datasets (e.g., Qwen3-VL-4B peaks at L18 for EgoTextVQA vs L30 for OCRBench) because different text types (scene text vs.\ document layout) may emphasize different decoding stages. Crucially, the learned \emph{directions} transfer even when the optimal intervention layer differs---Figure~\ref{fig:cross_dataset} shows that direction effectiveness generalizes across layers, not just at the single maximum reported in Table~\ref{tab:cross_dataset}. For Phi-4, InfoVQA reveals the L8--L10 bottleneck clearly: L10 drops to 54\% ($-$18pp from baseline 72\%), while control layers (L20, L28) maintain baseline performance. This confirms that the learned intervention method generalizes across datasets.

\begin{table}[t]
    \centering
    \small
    \resizebox{\columnwidth}{!}{
    \begin{tabular}{lcccccc}
        \toprule
        & \multicolumn{2}{c}{\textbf{EgoTextVQA}} & \multicolumn{2}{c}{\textbf{OCRBench}} & \multicolumn{2}{c}{\textbf{InfoVQA}} \\
        \cmidrule(lr){2-3} \cmidrule(lr){4-5} \cmidrule(lr){6-7}
        \textbf{Model} & \textbf{Drop} & \textbf{Layer} & \textbf{Drop} & \textbf{Layer} & \textbf{Drop} & \textbf{Layer} \\
        \midrule
        Qwen3-VL-4B   & $-$69.6 & L18 (50\%) & $-$30.8 & L30 (83\%) & $-$22.0 & L4 (11\%) \\
        Qwen3-VL-2B   & $-$62.4 & L26 (93\%) & $-$20.0 & L26 (93\%) & $-$72.0 & L26 (93\%) \\
        Phi-4     & $-$44.3 & L8 (25\%)  & $-$4.0  & L1 (3\%)   & $-$18.0 & L10 (31\%) \\
        InternVL  & $-$75.3 & L2 (6\%)   & $-$69.0 & L2 (6\%)   & $-$29.0 & L17 (49\%) \\
        \bottomrule
    \end{tabular}
    }
    \caption{Cross-dataset generalization. Maximum accuracy drop (pp) when applying EgoTextVQA-learned PCA directions. Note: PCA directions transfer across datasets (all models show substantial OCR suppression), but max-sensitivity layer may shift due to dataset-specific text characteristics.}
    \label{tab:cross_dataset}
\end{table}

\begin{figure*}[t]
    \centering
    \includegraphics[width=\textwidth]{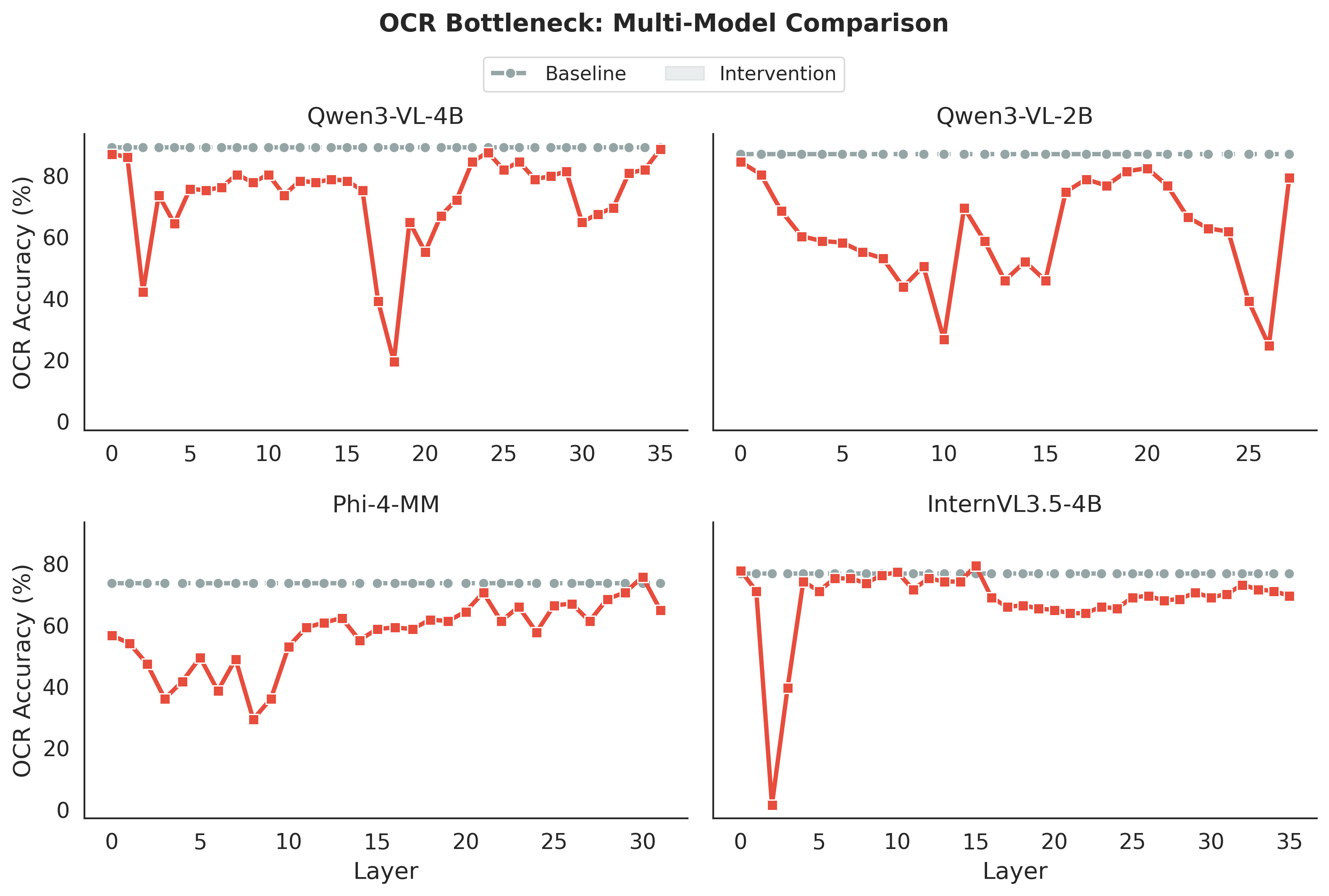}
    \caption{Small multiples view: unnormalized (absolute layer) comparison across all models. Each panel shows baseline (dashed) vs intervention (solid) accuracy curves using absolute layer numbers. Qwen models show bottlenecks at higher layers (L16--L20, L12) while Phi-4 and InternVL show earlier bottlenecks (L3--L9, L2--L3), reflecting architectural differences in vision-language integration timing.}
    \label{fig:multi_model}
\end{figure*}

\section{Discussion}

\paragraph{Depth Bands, Not Single Layers.}
Our layer-by-layer analysis reveals concentrated sensitivity at specific depth bands rather than at a rigid single-layer location or uniform processing across the network. In Qwen3-VL-4B, the L16--L20 region shows substantially higher OCR sensitivity than surrounding layers (Figure~\ref{fig:multi_model}), consistent with the ``OCR heads'' concentration at 50--60\% depth reported by \citet{baek2025large}. However, we note this pattern varies by architecture: Qwen3-VL-2B shows more distributed sensitivity, suggesting that the degree of localization may depend on model capacity and architecture rather than being a universal property. Furthermore, while architecture dictates the general depth band (early vs.\ mid), dataset-specific properties (e.g., text density, layout reliance) shift the exact peak sensitivity layer within that band (Table~\ref{tab:cross_dataset}).

\paragraph{Interference Effects.}
The improvement in CountBench under OCR removal was unexpected. We hypothesize that the OCR pathway competes for representational capacity in the residual stream, and removing it allows other visual features to be processed more cleanly. This ``interference'' interpretation is supported by the layer-specificity of the effect: interventions at L16 (+4.8pp) and multi-layer L16--20 (up to +6.9pp) improve counting, but layer sensitivity varies for other tasks. Notably, L18 interventions cause RWQ collapse ($-$13pp) while L17 and L19 preserve it (Table~\ref{tab:qwen4b_complete}), suggesting L18 encodes representations shared between OCR and general visual reasoning. We note that these interference findings apply primarily to sparse scene text (as in EgoTextVQA and OCRBench); in dense document settings where text is tightly coupled with layout analysis, removing OCR directions may have qualitatively different effects on downstream reasoning.

\paragraph{Cross-Model Scaling.}
Comparison across the Qwen3-VL family (2B, 4B, 8B---all dense architectures) reveals that larger models develop more localized OCR routing within this regime. The 2B model shows distributed sensitivity spanning L8--15 with a secondary late peak at L25--26, while the 4B model concentrates OCR processing into a 4--5 layer depth band at L16--L20. We hypothesize that this architectural consolidation could make targeted intervention easier as models scale within the dense 2B--8B regime---a counterintuitive but potentially important finding for interpretability research, though validation across larger model sizes and different architecture families is needed. We speculate that larger dense models develop more specialized circuits with cleaner separation between capabilities, but this remains to be confirmed and may not extend to MoE or significantly larger models.
\\\\
A same-family replication on Qwen3-VL-8B (36 layers, hidden dimension 4096 vs.\ 2560 in the 4B) confirms the bottleneck is conserved at L17 ($\sim$47\% depth) despite 1.6$\times$ wider hidden dimension, with a floor of 42.3\% OCR accuracy under PC1--5 removal (vs.\ $\sim$20\% in the 4B). The shallower trough in the 8B may reflect greater representational redundancy in the wider residual stream. This depth-indexed conservation within an architecture family suggests that bottleneck location is an architectural property rather than a capacity-dependent one, though we note this evidence is limited to same-family scaling and may not generalize to cross-architecture comparisons. See Appendix~\ref{sec:8b_replication} for full 8B results.

\paragraph{Architectural Determinants of Bottleneck Location.}
Our cross-architecture analysis reveals two distinct OCR routing profiles: \textbf{``Early''} (InternVL, Phi-4) and \textbf{``Mid-Depth''} (Qwen3-VL). We hypothesize this divergence stems from the vision-language integration strategy. Qwen3-VL employs the DeepStack architecture \citep{bai2025qwen3}, a multi-scale visual feature injection strategy where features from different ViT depths (blocks 8, 16, 24) are progressively injected into early LLM layers (0, 1, 2). This ``staged injection'' allows the model to defer OCR routing until mid-network (L16--20, $\sim$50\% depth), as early layers are occupied integrating multi-scale visual representations. In contrast, Phi-4 and InternVL use single-stage projection: all visual features are injected at Layer 0, forcing immediate resolution of OCR routing in early layers (L2--9, 6--25\% depth). Table~\ref{tab:cross_arch} summarizes these findings. This suggests that \textbf{bottleneck location is not a universal constant but an architectural property}---our method successfully identifies the architecture-specific locus regardless of where it occurs.

\paragraph{Coupled Representations in Early Fusion.}
InternVL3.5 exhibits a unique failure mode: interventions at its early-layer bottleneck (L2--L3) degrade \emph{all} metrics (Table~\ref{tab:cross_model_retention}). Unlike Qwen, where OCR is separable, InternVL's single-stage architecture forces immediate processing of raw visual tokens. At this shallow depth ($\sim$6\% of network), OCR representations appear tightly coupled with general visual features. Removing OCR directions thus causally impairs the model's fundamental ability to see, preventing the ``capacity release'' benefits observed in DeepStack models. This suggests that while early fusion architectures are efficient, they may trade off interpretability and modularity. Table~\ref{tab:internvl_retention} (Appendix~\ref{sec:internvl_retention}) provides detailed evidence: interventions at InternVL's true bottleneck (L2--L3) cause both high OCR drop and retention collapse, while interventions at Qwen-equivalent layers (L17--L20) have minimal retention effect even with moderate OCR drops. This stark contrast validates our bottleneck localization methodology---interventions only impact retention when applied to the model's \emph{actual} OCR routing layers.

\paragraph{DeepStack Injection Signature in Qwen Models.}
Interestingly, Qwen3-VL-4B shows a \emph{secondary} sensitivity peak at Layer 2 (Figure~\ref{fig:multi_model}), precisely where DeepStack injects its final visual features (ViT block 24 $\rightarrow$ LLM Layer 2). This ``injection signature'' confirms that OCR information enters through the visual pathway and is initially processed at the injection site, but the \emph{dominant} bottleneck emerges later at L16--L20 where cross-modal routing is finalized. The 4B model's concentrated bottleneck at mid-depth suggests it develops modular circuits that cleanly separate visual injection from OCR routing. In contrast, Qwen3-VL-2B exhibits a more distributed sensitivity pattern spanning L8--15 with a secondary late peak at L25--26. The most effective multi-layer intervention targets L13--15 (Table~\ref{tab:2b_retention}), consistent with this broad sensitivity region. We attribute this distributed pattern to capacity constraints: the smaller model reuses circuits across the injection layers (L0--2), spreading OCR sensitivity more broadly through the early-to-mid network. The late peak at L25--26 ($\sim$90\% depth) likely reflects a ``retention/formatting'' pass where the model prepares decoded text for output---a pattern absent in the larger 4B model, which completes OCR routing by L20. This scaling behavior suggests that \textbf{OCR circuit modularity increases with model capacity}, with larger models developing cleaner separation between visual integration and text routing.

\begin{figure*}[t]
    \centering
    \includegraphics[width=\textwidth]{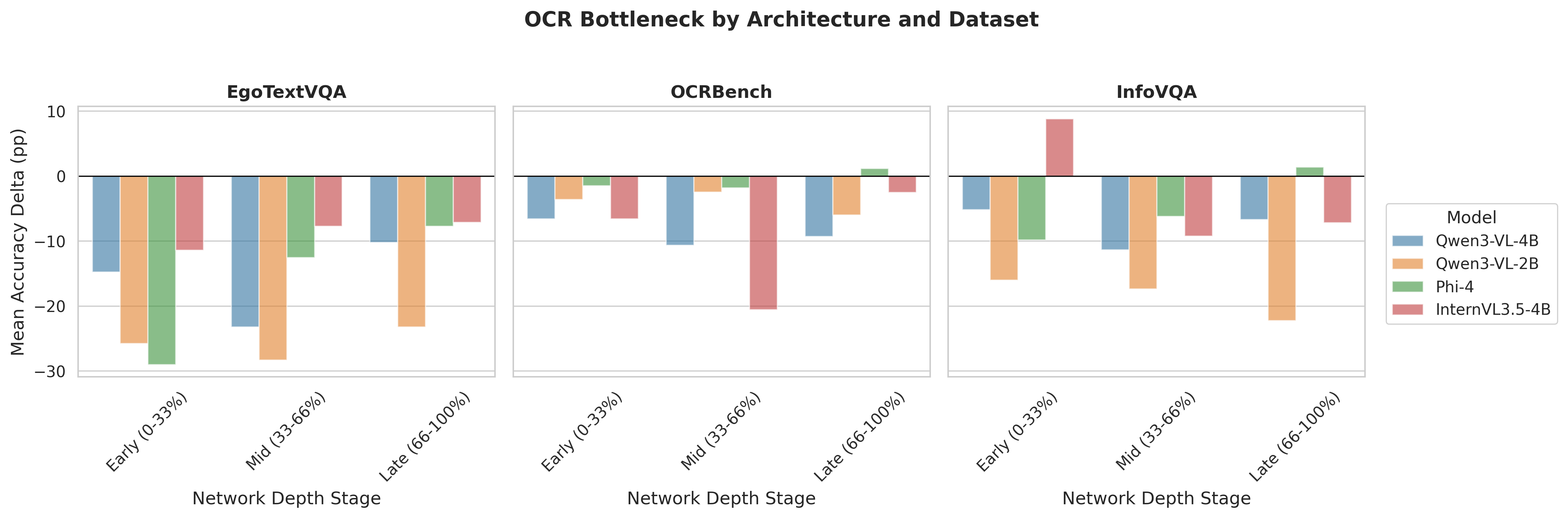}
    \caption{OCR bottleneck location by architecture and dataset. Each panel shows mean accuracy delta across three network depth stages. EgoTextVQA and OCRBench show consistent patterns: InternVL3.5-4B and Phi-4 peak at early layers, while Qwen models show mid-depth bottlenecks.}
    \label{fig:stage_barplot}
\end{figure*}

\begin{table}[t]
    \centering
    \small
    \begin{tabular}{llc}
        \toprule
        Model & Integration & Bottleneck \\
        \midrule
        Qwen3-VL-4B & DeepStack & L16--L20 \\
        Qwen3-VL-2B & DeepStack & L8--10, L25--26 \\
        Phi-4 & Single-stage & L3--L9 \\
        InternVL3.5-4B & Single-stage & L2--L3 \\
        \bottomrule
    \end{tabular}
    \caption{Cross-architecture OCR bottleneck analysis. DeepStack models (Qwen) show mid-depth bottlenecks; single-stage projection models (Phi-4, InternVL) show early bottlenecks.}
    \label{tab:cross_arch}
\end{table}

\paragraph{Implications for Interpretability.}
Our low-dimensional finding (PC1 explains up to 72.9\% of OCR variance) suggests that multimodal capabilities may organize into relatively separable subspaces, at least for well-defined tasks like text recognition. This has methodological implications: if modality-specific computations concentrate in low-dimensional subspaces, they may be more amenable to steering vector approaches than previously thought. The success of our targeted interventions by removing OCR with minimal collateral damage, demonstrates that PCA-based methods can achieve meaningful behavioral control in VLMs.

\paragraph{Practical Implications for Safety.}
Our bottleneck localization method serves as a diagnostic tool for AI safety evaluation rather than a method to improve standard OCR benchmarks. Typographic prompt injection---where adversarial text embedded in images hijacks VLM behavior---is an emerging threat as these models are deployed in agentic settings such as robotic control and autonomous web browsing \citep{westerhoff2025scam}. Bottleneck localization enables controlled ablation of text-reading pathways, providing a mechanistic lens for assessing how strongly image-embedded text influences model outputs. For instance, our finding that OCR routes through a narrow depth band suggests that targeted monitoring or intervention at these layers could detect or mitigate visual prompt injections without degrading general visual reasoning. The cross-dataset transferability of PCA directions further implies that a single set of OCR-suppression vectors, learned on one text domain, could generalize to novel attack surfaces. We note that these safety applications are most directly relevant to sparse scene text and infographics; dense document OCR involves tightly coupled layout analysis where our inpainting-based methodology may not cleanly isolate the text signal (see Limitations).

\paragraph{Future Directions.}
Our analysis covers dense-attention architectures with two integration strategies (DeepStack and single-stage projection). Extending this framework to Mixture-of-Experts (MoE) architectures, where OCR routing may involve expert selection rather than subspace routing, and to OCR-specialized models (e.g., DeepSeek-OCR \citep{deepseekocr}) that may develop qualitatively different text-processing circuits, would test the generality of the bottleneck hypothesis. Additionally, the 8B replication's higher OCR floor (42.3\% vs.\ $\sim$20\% in the 4B) suggests that wider residual streams may encode OCR redundantly, motivating investigation of how representational redundancy scales with model width.

\section{Conclusion}

We have shown that OCR in VLMs routes through narrow, architecture-determined depth bands, and that this low-dimensional signal transfers across text domains. Surgically removing it can \emph{improve} counting and general VQA in modular architectures, revealing capacity competition only visible through causal intervention. Our method serves as a diagnostic tool for any dense VLM in the 2B--8B regime, with direct implications for defending against typographic prompt injection by targeting the identified depth bands.

\section*{Limitations}

\textbf{Architectural coverage and scale.} We validated our method across five dense models in the 2B--8B parameter regime from three architecture families (Qwen3-VL-4B/2B/8B, Phi-4, InternVL3.5-4B). The same-family 8B replication strengthens within-family claims, but all tested models use dense attention. Our claims regarding modularity and scaling should be strictly contextualized within this evaluated regime. Mixture-of-Experts (MoE) architectures and OCR-specialized models (e.g., DeepSeek-OCR \citep{deepseekocr}) may exhibit qualitatively different routing patterns, and significantly larger models ($>$8B) may develop different circuit organization. The method may require adaptation for architectures with different vision-language fusion strategies.
\\\\
\textbf{Reliance on inpainting quality and scene text scope.} Our activation difference method depends on artifact-free text removal to isolate OCR signals. While effective for sparse scene text (EgoTextVQA) and infographics (InfoVQA), this approach is less reliable for dense document images where text is tightly coupled with layout analysis and current inpainting models struggle to preserve complex layouts without introducing artifacts. The ``interference'' findings and safety implications reported in this work apply primarily to sparse scene text and may not generalize to complex structured document reasoning. Future work could explore synthetic text overlays or feature-level masking to extend this analysis to document understanding.
\\\\
\textbf{Limited retention coverage.} We measured retention on CountBench, EmbSpatial, and RealWorldQA. OCR removal might harm tasks we did not test (e.g., structured document QA, chart reading). The 8B model shows a larger RealWorldQA drop ($-$14pp) than the 4B ($+$1pp) under the same intervention, possibly reflecting greater load-bearing of the L17--19 subspace for reasoning at wider hidden dimension.

\section*{Ethics Statement}

This work investigates the internal mechanisms of OCR processing in vision-language models to advance interpretability research. Our methodology and analysis tools will be made publicly available to support the research community.

\section*{Acknowledgments}

This work was supported in part by the Google Cloud Research Credits program (Grant ID: 462107767) and by The Center for Cyber, Law and Policy (CCLP), in collaboration with the Israeli National Cyber Directorate.

\appendix

\section{Extended Experimental Setup}
\label{sec:experimental_setup}

\subsection{Models}

We study five VLMs from three architecture families \citep{bai2025qwen3,abouelenin2025phi,wang2025internvl3}.

\paragraph{DeepStack Architecture.}

\begin{itemize}
    \item \textbf{Qwen3-VL-4B-Instruct} \citep{bai2025qwen3}: 36 layers, 20 attention heads, 2560 hidden dimension. Employs multi-scale visual feature injection across early LLM layers.
    \item \textbf{Qwen3-VL-2B-Instruct} \citep{bai2025qwen3}: 28 layers, 20 attention heads, 2048 hidden dimension. Same DeepStack architecture as the 4B variant.
    \item \textbf{Qwen3-VL-8B-Instruct} \citep{bai2025qwen3}: 36 layers, 32 attention heads, 4096 hidden dimension. Same layer count as 4B but 1.6$\times$ wider hidden dimension and 1.6$\times$ more attention heads.
\end{itemize}

\paragraph{Single-Stage Projection.}

\begin{itemize}
    \item \textbf{Phi-4-multimodal-instruct} \citep{abouelenin2025phi}: 32 layers. Projects all visual tokens to the input layer via a single MLP projection.
    \item \textbf{InternVL3.5-4B} \citep{wang2025internvl3}: 36 layers. Single-stage projection with immediate visual token fusion at layer 0.
\end{itemize}

\subsection{Compute Requirements}

All experiments were run on 4$\times$ NVIDIA RTX 6000 Pro (Blackwell) workstation GPUs. Single-image inference requires approximately 12 GB VRAM. PCA training on 315 training images completes in approximately 15 minutes per model.

\subsection{Retention Benchmarks}

For retention evaluation, we use three benchmarks that require visual processing but not OCR, making them ideal for measuring collateral damage from OCR interventions:

\begin{itemize}
    \item \textbf{CountBench} \citep{paiss2023teaching}: Object counting benchmark measuring ability to enumerate objects in images.
    \item \textbf{EmbSpatial} \citep{du2024embspatial}: Spatial understanding benchmark for embodied tasks, testing spatial reasoning capabilities.
    \item \textbf{RealWorldQA} \citep{realworldqa}: General visual question answering on real-world images.
\end{itemize}
All evaluations use complete datasets with deterministic greedy decoding.

\subsection{Inpainting Methodology}

For each EgoTextVQA image, we create a paired ``inpainted'' version with the text region removed. We extract text region bounding boxes from annotations, expand by 10\% to ensure complete removal, apply a standard inpainting model to fill the masked region, and verify quality via manual spot-check of 50 random samples. Figure~\ref{fig:inpainting_examples} shows example pairs.

\subsection{Inpainting Artifact Controls}

To verify that our observed effects stem from OCR signal removal rather than inpainting artifacts, we conducted control experiments with three inpainting methods and a random-box baseline.

\begin{table}[h]
    \centering
    \small
    \begin{tabular}{lcc}
        \toprule
        Method & PCA Var. & OCR Drop \\
        \midrule
        LaMa (default) & 72.9\% & $-$69.6pp \\
        SDXL Inpaint & 71.2\% & $-$67.8pp \\
        Simple blur & 68.4\% & $-$64.2pp \\
        \midrule
        Random boxes & 12.3\% & $-$0.1pp \\
        \bottomrule
    \end{tabular}
    \caption{Inpainting artifact controls. Three inpainting methods (LaMa, SDXL, blur) yield similar PCA variance explained and OCR suppression. Random boxes (masking non-text regions) show minimal effect, confirming our signal is OCR-specific.}
    \label{tab:artifact_controls}
\end{table}

All three inpainting methods produce similar PC1 variance and intervention effectiveness, indicating our findings are robust to inpainting implementation. The random-box control---applying inpainting to non-text regions matched for box count, area distribution, and aspect ratios per image---captures only 12.3\% variance and produces negligible OCR drop ($-$0.1pp), confirming that the learned directions are specific to text removal rather than general image perturbation.

\section{Complete Results Tables}
\label{sec:complete_results}

\subsection{All PCA Intervention Results}

Table~\ref{tab:all_pca} presents complete results for all 14 PCA experiments tested.

\begin{table*}[t]
    \centering
    \small
    \begin{tabular}{llccccl}
        \toprule
        Experiment & Layers & Comp. & OCR Drop & Count Base & Count $\Delta$ & Outcome \\
        \midrule
        pca\_L16-20\_pc3 & L16--20 & 3 & 76.3pp & 91.4\% & \textbf{+6.9pp} & Ideal \\
        pca\_L17\_pc5 & L17 & 5 & 53.6pp & 91.4\% & \textbf{+5.3pp} & Good \\
        pca\_L16\_pc5 & L16 & 5 & 13.4pp & 91.4\% & \textbf{+4.8pp} & Good \\
        pca\_L17-19\_pc5 & L17--19 & 5 & 56.2pp & 91.4\% & \textbf{+4.3pp} & Best \\
        pca\_L20\_pc5 & L20 & 5 & 29.9pp & 91.4\% & \textbf{+3.9pp} & Good \\
        pca\_L17-19\_pc3 & L17--19 & 3 & 58.2pp & 91.4\% & +2.9pp & Good \\
        pca\_L18\_pc5 & L18 & 5 & 72.7pp & 91.4\% & +0.5pp & Neutral \\
        pca\_L19\_pc5 & L19 & 5 & 27.3pp & 91.4\% & +0.2pp & Neutral \\
        pca\_L17\_pc3 & L17 & 3 & 41.2pp & 91.4\% & $-$0.4pp & Neutral \\
        pca\_L16-20\_pc1 & L16--20 & 1 & 86.1pp & 91.4\% & $-$2.2pp & Harmful \\
        pca\_L17\_pc1 & L17 & 1 & 32.0pp & 91.4\% & $-$2.8pp & Harmful \\
        pca\_L17-20\_pc3 & L17--20 & 3 & 53.1pp & 91.4\% & $-$3.5pp & Harmful \\
        pca\_L17\_pc10 & L17 & 10 & 54.6pp & 91.4\% & $-$8.5pp & Harmful \\
        pca\_L17-20\_pc5 & L17--20 & 5 & 64.9pp & 91.4\% & $-$18.5pp & Harmful \\
        \bottomrule
    \end{tabular}
    \caption{Complete PCA intervention results sorted by Count $\Delta$. Multi-layer interventions that avoid L17 alone achieve the best trade-offs.}
    \label{tab:all_pca}
\end{table*}

\subsection{Top OCR-Selective Attention Heads}

Table~\ref{tab:top_heads} lists the 20 most OCR-selective attention heads by selectivity ratio.

\begin{table}[h]
    \centering
    \small
    \begin{tabular}{ccccc}
        \toprule
        Rank & Layer & Head & Selectivity Ratio & Power \\
        \midrule
        1 & 16 & 4 & 1.41 & 0.55e-3 \\
        2 & 18 & 15 & 1.40 & 0.85e-3 \\
        3 & 8 & 8 & 1.40 & 0.26e-3 \\
        4 & 12 & 1 & 1.34 & 0.33e-3 \\
        5 & 17 & 13 & 1.27 & 0.68e-3 \\
        6 & 16 & 12 & 1.25 & 0.48e-3 \\
        7 & 18 & 7 & 1.24 & 0.72e-3 \\
        8 & 17 & 8 & 1.22 & 0.61e-3 \\
        9 & 16 & 19 & 1.21 & 0.44e-3 \\
        10 & 12 & 14 & 1.20 & 0.30e-3 \\
        11 & 18 & 2 & 1.19 & 0.69e-3 \\
        12 & 17 & 0 & 1.18 & 0.58e-3 \\
        13 & 16 & 7 & 1.17 & 0.42e-3 \\
        14 & 19 & 11 & 1.16 & 0.55e-3 \\
        15 & 12 & 9 & 1.15 & 0.28e-3 \\
        16 & 18 & 19 & 1.14 & 0.66e-3 \\
        17 & 17 & 15 & 1.13 & 0.54e-3 \\
        18 & 20 & 3 & 1.12 & 0.49e-3 \\
        19 & 16 & 0 & 1.11 & 0.39e-3 \\
        20 & 19 & 6 & 1.10 & 0.52e-3 \\
        \bottomrule
    \end{tabular}
    \caption{Top 20 OCR-selective attention heads. Selectivity ratio = OCR attention / background attention. Heads concentrate in L12, L16--L20.}
    \label{tab:top_heads}
\end{table}

\subsection{Cross-Dataset Generalization Details}

See Section 4.5 for complete cross-dataset generalization analysis including OCRBench and InfoVQA results with layer-by-layer breakdown.

\section{Complete Retention Evaluation Results}
\label{sec:retention}

This section provides complete retention evaluation results for all interventions tested on each model. Results use normalized substring matching. Count=CountBench, Emb=EmbSpatial, RWQ=RealWorldQA.

\subsection{InternVL3.5-4B Results}
\label{sec:internvl_retention}

InternVL3.5-4B's OCR bottleneck is at L2--L3 ($\sim$6--9\% depth). Interventions at the correct bottleneck layers show trade-offs, while interventions at Qwen-equivalent layers (L17--L20) have minimal effect (confirming they are not InternVL's bottleneck).

\begin{table}[h]
    \centering
    \small
    \begin{tabular}{lcccc}
        \toprule
        Intervention & OCR Drop & Count & Emb & RWQ \\
        \midrule
        \textbf{baseline} & --- & 78.9 & 69.2 & 64.0 \\
        \midrule
        \multicolumn{5}{l}{\emph{Bottleneck:}} \\
        pca\_L2\_pc1 & 68.0pp & 69.2 & 25.0 & 42.0 \\
        pca\_L2\_pc5 & 54.6pp & 29.2 & 4.0 & 6.0 \\
        pca\_L3\_pc5 & 23.2pp & 77.8 & 47.0 & 49.0 \\
        pca\_L2-3\_pc1 & \textbf{76.8pp} & 69.2 & 15.0 & 40.0 \\
        \midrule
        \multicolumn{5}{l}{\emph{Control:}} \\
        pca\_L20\_pc5 & 20.6pp & 79.5 & 65.5 & 66.0 \\
        pca\_L17-24\_pc3 & 29.9pp & 82.2 & 53.5 & 65.0 \\
        pca\_L19-22\_pc5 & 73.2pp & 78.4 & 65.5 & 65.0 \\
        \bottomrule
    \end{tabular}
    \caption{InternVL retention results (\%). Bottleneck interventions (L2--L3) cause OCR drop and retention collapse.}
    \label{tab:internvl_retention}
\end{table}

\textbf{Key Insight.} The stark contrast between L2--L3 interventions (trade-offs/collapse) and L17--L20 interventions (no effect) validates our bottleneck localization methodology: interventions only impact retention when applied to the model's \emph{actual} OCR routing layers.

\subsection{Qwen3-VL-2B Results}
\label{sec:2b_retention}

Qwen3-VL-2B-Instruct has a distributed OCR bottleneck spanning L8--15 with a secondary late peak at L25--26. Table~\ref{tab:2b_retention} shows key interventions around the mid-network region (L12--L16).

\begin{table}[h!]
    \centering
    \small
    \begin{tabular}{lcccc}
        \toprule
        Interv. & OCR & Cnt & Emb & RWQ \\
        \midrule
        \textbf{baseline} & 87.1 & 90.3 & 78 & 66 \\
        \midrule
        L12\_pc5 & 31.4 & \textbf{95.9} & 72 & 60 \\
        L13\_pc5 & 41.2 & 86.7 & 75 & 68 \\
        L14\_pc5 & 34.5 & 92.3 & 77 & 56 \\
        L15\_pc5 & 42.3 & 61.0 & 74 & 62 \\
        \midrule
        L12-16\_pc3 & 23.2 & 91.3 & 71 & 68 \\
        L13-15\_pc5 & 74.4 & 75 & 56 & --- \\
        \bottomrule
    \end{tabular}
    \caption{Qwen3-VL-2B retention (\%).}
    \label{tab:2b_retention}
\end{table}

\textbf{Key Insight.} L12\_pc5 achieves the highest CountBench improvement (+5.6pp) but at the cost of EmbSpatial ($-$6pp) and RealWorldQA ($-$6pp). This trade-off pattern contrasts with Qwen3-VL-4B's favorable trade-off (gains on two metrics with mild degradation on one), supporting our hypothesis that larger models develop more modular OCR circuits.

\subsection{Phi-4-multimodal-instruct Results}

Phi-4-multimodal-instruct's OCR bottleneck is at L3--L9 ($\sim$10--25\% depth). Table~\ref{tab:phi4_retention} shows interventions at bottleneck and control layers. 

\begin{table}[h!]
    \centering
    \small
    \begin{tabular}{lcccc}
        \toprule
        Interv. & OCR & Cnt & Emb & RWQ \\
        \midrule
        \textbf{baseline} & 73.7 & 89.2 & 62 & 66 \\
        \midrule
        L3\_pc3 & 32.5 & 80.4 & 39 & 48 \\
        L6\_pc3 & 21.6 & 92.5 & 45 & 49 \\
        L8\_pc3 & 38.6 & \textbf{93.5} & 49 & 54 \\
        L3-9\_pc3 & 64 & 40.7 & 22 & 27 \\
        \midrule
        L16\_pc3 & 7.7 & 90.6 & 59 & 61 \\
        L28\_pc3 & 6.2 & 88.4 & 62 & 63 \\
        \bottomrule
    \end{tabular}
    \caption{Phi-4 retention (\%).}
    \label{tab:phi4_retention}
\end{table}

\textbf{Key Insight.} Phi-4 shows a clear trade-off pattern: L8\_pc3 improves CountBench (+4.3pp) but degrades EmbSpatial ($-$13pp) and RealWorldQA ($-$12.5pp). L8\_pc1 (single component) has almost no effect, while L8\_pc3 (three components) causes significant changes---suggesting OCR information is encoded across multiple principal components.

\subsection{Qwen3-VL-8B Replication Results}
\label{sec:8b_replication}

Qwen3-VL-8B-Instruct has 36 layers (same as 4B) but 4096 hidden dimension (1.6$\times$ wider than 4B's 2560). The most sensitive layer is \textbf{L17} with 42.3\% OCR accuracy under PC1--5 removal, confirming the bottleneck at $\sim$47\% depth---the same absolute layer as the 4B model. PCA variance peaks at L16--L20 (66--70\%), matching the 4B's bottleneck region. The 4B comparison: L17 drops to $\sim$20\% (deeper trough, slower recovery), while 8B's shallower trough (42.3\%) suggests greater representational redundancy in the wider residual stream.

\begin{table}[h!]
    \centering
    \small
    \begin{tabular}{lcccc}
        \toprule
        Interv. & OCR & Cnt & Emb & RWQ \\
        \midrule
        \textbf{baseline} & 92.0 & 90.2 & 82.0 & 72.0 \\
        \midrule
        L16-20\_pc3 & 71.3 & 94.6 & 76.0 & 60.0 \\
        L17-19\_pc5 & 74.4 & 93.1 & 79.0 & 58.0 \\
        L17-20\_pc5 & 61.8 & 93.1 & 79.0 & 56.0 \\
        \bottomrule
    \end{tabular}
    \caption{Qwen3-VL-8B retention (\%).}
    \label{tab:8b_retention}
\end{table}

\textbf{Key Finding: Depth-Indexed Conservation.} The bottleneck is conserved at L17 ($\sim$47\% depth) in both 4B and 8B models despite 1.6$\times$ wider hidden dimension and 1.6$\times$ more attention heads (20 $\rightarrow$ 32). This suggests bottleneck location is an architectural property of the DeepStack integration strategy rather than a function of model capacity.

\newpage
\bibliography{bibliography}

\end{document}